\documentclass[11pt]{article}
\usepackage[preprint]{acl}

\usepackage{times}
\usepackage{latexsym}
\usepackage[T1]{fontenc}
\usepackage[utf8]{inputenc}
\usepackage{microtype}
\usepackage{inconsolata}
\usepackage{graphicx}
\usepackage{amsmath,amssymb}
\usepackage{booktabs}
\usepackage{xcolor}
\usepackage{xspace}

\title{\textsc{SpecMem}: Accelerating Agentic Tool Calling via Live Memory Management}

\author{
\textbf{Zihe Song}\textsuperscript{1} \quad
\textbf{Wenting Zhao}\textsuperscript{2} \quad
\textbf{Ye Liu}\textsuperscript{2} \quad
\textbf{Yuanjie Zhu}\textsuperscript{1} \\
\textbf{Ke Xu}\textsuperscript{1} \quad
\textbf{Shanghao Li}\textsuperscript{1} \quad
\textbf{Philip S. Yu}\textsuperscript{1} \\
\\
\textsuperscript{1}Department of Computer Science, University of Illinois Chicago \\
\textsuperscript{2}Salesforce Research
}

\newcommand{\method}{\textsc{SpecMem}\xspace}

\begin{document}
\maketitle

\begin{abstract}
Tool-calling agents are mostly optimized for accuracy; in deployment, however, their
growing cost is latency: every tool call is decoded autoregressively. Speculative
decoding accelerates this without changing the output, and retrieved past calls make
strong drafts under real use, where each user's behavior tends to repeat. Yet existing
systems keep no cross-session memory, or build the draft store once and never update
it. We propose \method, a tool-call memory kept alive and managed
throughout deployment: verified calls are written back online into bounded per-user
stores with LRU eviction, and a side-effect gate defers irreversible calls until
verification confirms. On multi-session, multi-user query streams served by three
open-weight models, a frozen store steadily loses acceptance while \method\ holds it: a
$27\%$ gap in mean accepted tokens ($+11\%$ to $+34\%$ across benchmarks, tracking how
much novel work arrives after the store is built) and up to $1.74\times$ wall-clock
speedup. Ablations attribute the gain almost entirely to keeping the store live; local
per-user scoping adds no acceptance but closes a measured cross-user leakage channel,
and gating on declared side effects, rather than on retrieval confidence, is what
eliminates irreversible mistakes.
\end{abstract}

\section{Introduction}

LLM agents increasingly act through tool calls: querying APIs, searching, and operating
software on a user's behalf \citep{qin2023toolllm,bfcl2024}. Most work on such agents
targets \emph{accuracy}: choosing the right tool with the right arguments
\citep{li2023apibank,tang2023toolalpaca}. In deployment, however, the cost that grows is
\emph{latency}: every call is decoded autoregressively, and an agent that pauses between
steps is impractical for interactive use, however accurate it is. This paper is about
making tool-calling agents faster without affecting what they output.

\begin{figure*}[t]
\centering
\includegraphics[width=0.97\textwidth]{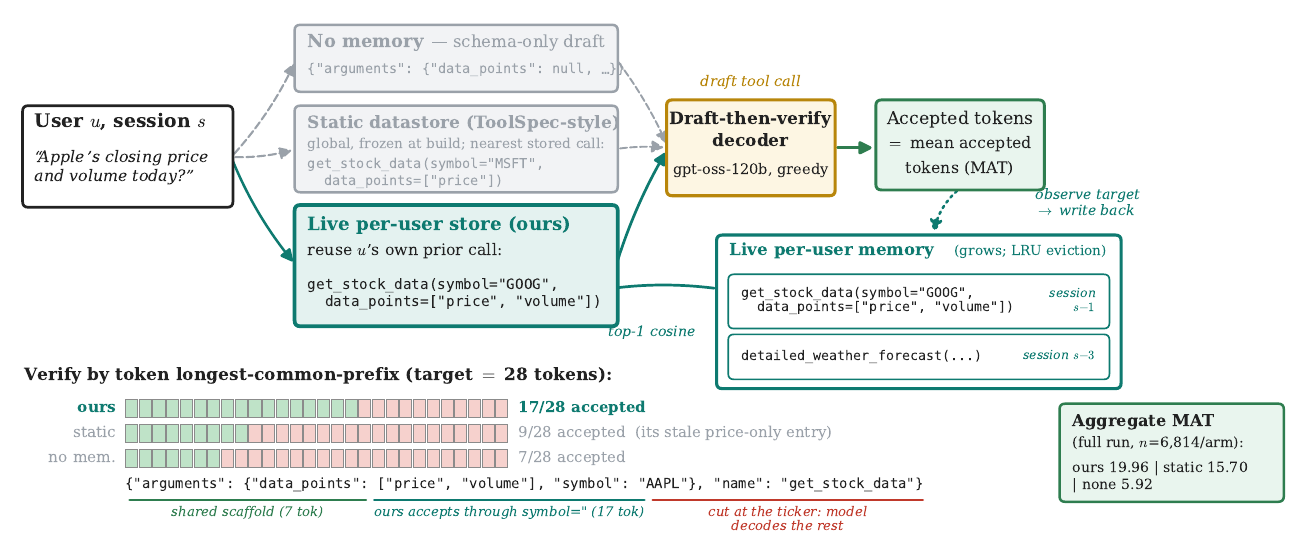}
\caption{A frozen \textsc{ToolSpec}-style datastore vs.\ \method\ on a recurring call:
user $u$ checked Google's stock the day before and now asks for Apple's. Each arm drafts
a candidate call for a draft-then-verify decoder over the fixed served model (acceptance
$=$ token longest-common-prefix). \method\ retrieves $u$'s own \texttt{GOOG} call,
accepted up to the changed ticker ($17$ of $28$ target tokens) with only the remainder
decoded; the frozen \textsc{ToolSpec}-style store's stale build-time entry accepts $9$,
the schema-only draft $7$ (full-run MAT $19.96$ vs.\ $15.70$ and $5.92$). Only the live
store writes the verified target back, so its advantage grows across sessions.}
\label{fig:arch}
\end{figure*}

A growing line of work targets exactly this efficiency. Speculative decoding lets a
cheap draft be verified by the target model, so the output is unchanged and the speedup
rests on how often the draft is right
\citep{leviathan2023speculative,chen2023accelerating}. Moving from general decoding to
agentic flows, speculation has been applied to the next tool call, its arguments, or the
following observation
\citep{sui2026paste,saberi2026spechop,hooper2026specagents,chung2026agentx}, and to
repeated output patterns cached across requests \citep{oliaro2024suffixdecoding}. Tool
calls are highly templated, so among draft sources, a similar past call retrieved from
\emph{memory} is especially likely to be right.

The scenario that matters in practice is tool calling under real use: an agent serves
many users over a long horizon, and each user's behavior tends to repeat, asking for
the weather, checking the same stock, or re-running yesterday's query with today's date.
Whether this recurrence can make agentic tool calling faster is the problem we focus
on. The closest work is \textsc{ToolSpec} \citep{xia2026toolspec},
which retrieves similar past calls from a datastore as drafts and reports large
speedups. Its store, however, is built once and then frozen (never updating, never
evicting, and shared globally across users), so the recurrence a returning user keeps
supplying is never reused (Figure~\ref{fig:arch}). A gap therefore remains between
static stores and an agent that stays efficient in this evolving scenario.

To close this gap, we propose \method, a tool-call memory that is kept \emph{alive} and
managed throughout deployment. Managing such a memory raises three coupled problems.
Because calls accumulate across sessions, the store must stay bounded yet fresh as the
user behavior drifts (\emph{lifecycle}). Because the agent serves many users, retrieval must
draw either on one shared global store or on each user's own history, a choice that
decides whose calls can appear in whose drafts (\emph{scope}). And because tools act on
the world, a draft executed before verification can commit an irreversible action, such
as sending an email (\emph{side effects}; safety here in the systems sense, not content
moderation). \method\ answers all three with online write-back into bounded, locally
scoped stores under LRU eviction, and a \emph{side-effect gate} that defers execution
of any call not declared safe until verification confirms (Figure~\ref{fig:arch}).
Overall, \method\ delivers up to $1.74\times$ wall-clock decode speedup
over vanilla decoding and a $+27\%$ post-warmup acceptance gain over the frozen store,
widening to $+46\%$ by the final session.

Our experiments yield three findings. First, \emph{liveness} drives acceptance: on
multi-session, multi-user query streams served by real open-weight models, the frozen store
degrades across sessions while the live store holds, an ablation attributes essentially
the entire gain to online write-back, and the gap scales with the share of novel calls
arriving after the store is built ($+11\%$ to $+27\%$ across $\tau^2$-bench domains
\citep{barres2025tau2}, $+34\%$ on Seal-Tools \citep{wu2024sealtools}). Second, local
scoping adds almost nothing to acceptance but closes a measured leakage channel: under a
global store, $14$--$68\%$ of a user's drafts came from \emph{other} users' histories.
Third, retrieval confidence cannot make speculative execution safe, since a stale
draft's similarity can be indistinguishable from a correct one's, whereas gating on declared
side effects removes irreversible mistakes by construction.

In sum, this paper (i) formulates persistent tool-call memory as a lifecycle problem
(write-back, partitioning, eviction, execution gating), instantiated in \method\
(\S\ref{sec:method}); (ii) shows that
a frozen store degrades while a live one holds, quantifies how the gap scales with
post-freeze novelty, and attributes the gain to liveness rather than to store scope,
eviction, or the choice of retriever, while measuring the cross-user draft leakage a
global store incurs (\S\ref{sec:exp}--\S\ref{sec:analysis}); (iii) identifies speculative execution of
side-effecting tools as a hazard that confidence thresholds do not remove, and proposes a
schema-level side-effect gate that does, defined on declared side effects rather than
idempotency (\S\ref{sec:safety}); and (iv) validates the findings across three model
families and six benchmarks (BFCL, Seal-Tools, live multi-turn $\tau^2$-bench traces,
API-Bank, ToolAlpaca, and ToolBench), including real wall-clock measurement
(\S\ref{sec:maintable}--\S\ref{sec:sealtools}).

\section{Related Work}
\label{sec:related}

\paragraph{Speculative decoding for tool-calling agents.}
Draft-then-verify decoding accelerates inference without changing the output distribution
\citep{leviathan2023speculative,chen2023accelerating}; drafts can come from self-drafting
heads \citep{cai2024medusa,li2024eagle,li2024eagle2}, parallel decoding
\citep{fu2024lookahead}, or, most relevant here, from \emph{retrieval}: REST verifies
recurring $n$-grams from a datastore \citep{he2024rest} and \textsc{SuffixDecoding} grows an online
suffix cache \citep{oliaro2024suffixdecoding}, which we adopt as a live token-match baseline
(\S\ref{sec:ablation}); see \citet{xia2024specbench} for a survey. Retrieval
drafting suits agents because tool calls are highly \emph{templated} and recurrent
\citep{yao2022react,schick2023toolformer,qin2023toolllm,liu2024toolace}, the repetition a
memory exploits. Closest in spirit, \emph{online speculative decoding} \citep{liu2024online}
fine-tunes a neural drafter under distribution shift; we ask the same liveness question for
\emph{retrieval} stores, where staying live is a write-back, not a training loop. A
concentrated 2024--2026 wave then speculates over agent actions: the next tool call
\citep{sui2026paste}, tool observations \citep{saberi2026spechop}, on-device
\citep{chung2026agentx}, concurrent I/O overlap \citep{hooper2026specagents}, and forked,
beam, and multimodal variants \citep{bai2026spork,song2026bpaste,huang2026speceyes}. But
\emph{none keep a persistent cross-session memory}: each speculates from within-task context
and discards it.

\paragraph{The closest prior datastore, and its gap.}
The closest prior system to ours is \textsc{ToolSpec} \citep{xia2026toolspec}, which
maintains a historical tool-call datastore retrieved by embedding similarity with a
schema-aware FSM and reports $3.5$--$4.2\times$ speedups. Its evaluation store, however,
is fixed and globally shared, with no online write-back, eviction, or gating of
side-effecting tools; its appendix explicitly leaves memory management to future work. \method\ targets
exactly this gap, treating the store as a lifecycle and evaluating the cross-session
dynamics that a fixed store cannot sustain.

\paragraph{Agent memory, caching, and tool safety.}
Agent memory systems keep bounded, updated, user-scoped stores (MemGPT
\citep{packer2023memgpt}, MemoryBank \citep{zhong2023memorybank}, Generative Agents
\citep{park2023generative}, Reflexion \citep{shinn2023reflexion}, A-MEM
\citep{xu2025amem}, Agent Workflow Memory \citep{wang2024awm}), but for knowledge or
skill recall, not speculative drafts; semantic caches like GPTCache
\citep{bang2023gptcache} evict for \emph{response reuse}, where a stale hit costs
correctness rather than acceptance. On tool-use safety, ToolEmu \citep{ruan2023toolemu}, R-Judge
\citep{yuan2024rjudge}, ToolSword \citep{ye2024toolsword}, and GuardAgent
\citep{xiang2024guardagent} document that tool-use risk is real and \emph{graded}, and
speculative-execution engines buffer side effects until commit
\citep{nightingale2005speculative}. Yet the systems above uniformly assume tools are safe to execute speculatively (Ghost Tool Calls \citep{mohammadi2026ghost} gates \emph{privacy}, a complementary
axis); this motivates our severity-weighted cost of being wrong and the hard gate
(\S\ref{sec:safety}).

\section{Methodology}
\label{sec:method}

Our methodology consists of three components: (1)~\emph{data construction}
extending single-shot tool-calling benchmarks into the multi-user, multi-session
scenario we target (\S\ref{sec:method-workload}); (2)~\method, a tool-call memory
maintained as a managed lifecycle behind a speculative decoder
(\S\ref{sec:method-specmem}); and (3)~a \emph{side-effect gate} governing when a drafted
call may execute before verification (\S\ref{sec:method-safety}).

\subsection{Preliminaries}
\label{sec:method-prelim}

Speculative decoding lets a cheap source propose a draft that the target model verifies
in one forward pass, accepting the longest correct token prefix and re-decoding from the
first mismatch \citep{leviathan2023speculative,chen2023accelerating}; the output is
unchanged, so the speedup depends only on how much of the draft is accepted. In tool
calling the draft is a candidate call, and a similar past call retrieved from a
datastore makes a strong one \citep{xia2026toolspec}. Our setting is a deployment in
which a fixed served model serves many users over many sessions; the object of study is
the datastore: how it is scoped, updated, bounded, and gated as user behavior evolves.

\subsection{Extending benchmarks to the deployment scenario}
\label{sec:method-workload}

Public tool-calling benchmarks are single-shot pools of one-off tasks with no notion of
users or time, so we extend them into multi-user, multi-session query streams. Benchmark
tasks are partitioned across $U$ simulated users; each user's tasks are then re-issued
as queries over ordered sessions, and every recurrence perturbs the task's numeric
argument values, so a user's queries repeat in structure but not verbatim. Session~0 is
a cold start for every arm, and part of each user's tasks first appear only in later
sessions, so novel work keeps arriving after warmup. The construction is benchmark-agnostic (we instantiate it on
BFCL and Seal-Tools, \S\ref{sec:setup}) and identical across arms, so it cannot favor
any one of them (synthetic structure: \S\ref{sec:limitations}).

\subsection{\method: live memory management}
\label{sec:method-specmem}

Given this scenario, \method\ maintains the draft store as a managed lifecycle rather
than a static index (Figure~\ref{fig:arch}).

\paragraph{Retrieval-based drafting.}
Each user owns a store: an ordered map keyed by user id, holding
$(\text{embed}(\text{query}), \text{canonical\_call})$ pairs from that user's history.
At draft time, \method\ embeds the incoming query and retrieves the user's nearest entry
by cosine similarity; if the similarity clears $\tau{=}0.35$, the stored call is
returned as the draft and promoted to most recently used. If the user's store is empty
or too dissimilar (the cold-start case), \method\ backs off to a small \emph{shared}
store seeded from population history, and finally to a schema-only skeleton (tool name
plus argument placeholders), whose tokens any correct call to that schema must share as
a prefix.

\paragraph{Online write-back and bounded eviction.}
After every verified call, \method\ inserts the confirmed call, keyed by its query
embedding, into the issuing user's store. Write-back is unconditional: the target's
output is produced in full regardless of how much of the draft was accepted, and the
confirmed call is the freshest evidence of the user's current behavior. Each user's
store is bounded at capacity $C$ ($48$ in full-scale runs); on overflow the least
recently used entry is evicted (LFU is a drop-in alternative). Store scope (local
per-user keying vs.\ global) and lifecycle (bounded and evicting) are thus the variables
that distinguish \method\ from a frozen global store.

\subsection{Gating speculative execution of side-effecting tools}
\label{sec:method-safety}

Speculative \emph{execution} hides tool latency by running the drafted tool before
verification completes. For a read-only tool a wrong speculation wastes work; for a tool
with externally visible side effects (\texttt{transfer\_funds}, \texttt{drop\_table},
\dots) it commits harm that verification then rejects too late. We emphasize that the
relevant property is \emph{not} idempotency: an idempotent operation can still be harmful
(deleting a file twice yields the same state as deleting it once, but a wrongly drafted
deletion is still a loss), and a nominally non-idempotent call can be safe inside a
transaction or dry run. The property that matters is whether executing the call before
verification can commit an unrecoverable, externally visible effect. We therefore tag
each tool with a \texttt{safe\_to\_speculate} flag as part of its declared schema: true
for read-only or side-effect-free calls and for calls whose effects are transactionally
buffered or compensatable, false for irreversible side-effecting calls, whether
idempotent or not. We then apply a schema-level \textbf{side-effect gate}: only
flagged-safe drafts may be speculatively executed; all others are still drafted and
verified speculatively, but execution is deferred until verification confirms.
Call-dependent cases (e.g.\ a payment that is safe only when an idempotency key is
supplied) need a per-call ternary gate, which strictly dominates the binary flag
(\S\ref{sec:limitations}).

\paragraph{Cost of being wrong.}
A raw count treats moving money and sending a duplicate message alike, so we score each
unsafe tool with a severity $\sigma(t)\in[1,10]$ ($\sigma{=}0$ if safe to speculate). A
policy's \textbf{cost of being wrong} is $\sum \sigma(t)$ over its wrong irreversible
speculative executions. We compare \emph{naive} (execute every accepted draft), a
\emph{confidence gate} (execute an unsafe draft only above a similarity threshold
$\tau_c$, plausible but insufficient), and our \emph{side-effect gate}. The gate defers
only execution, not drafting: speculation on safe calls is unaffected; the cost is
forgoing speculative execution on unsafe drafts, including correct ones (quantified in
\S\ref{sec:safety}).

\section{Experiments}
\label{sec:exp}

\subsection{Experimental setup}
\label{sec:setup}

\paragraph{Data.} We instantiate the construction of \S\ref{sec:method-workload}
on the BFCL~v4 simple/multiple/parallel pool \citep{bfcl2024}. Each of $40$ simulated
users is assigned $15$ benchmark tasks, disjoint across users. Session~0 (warmup) issues
$8$ of each user's tasks once; the remaining $7$ first appear spread across
sessions~1--11, so novel work keeps arriving after warmup. Each of sessions~1--11 issues
$6$ queries per user, drawn from the tasks seen so far and weighted toward recently
arrived ones, with numeric argument values perturbed on every recurrence. This yields
$2{,}960$ queries per seed ($8{,}880$ over $3$ seeds); queries the model declines are
excluded from all arms symmetrically, leaving $6{,}814$ scored post-warmup queries per
arm (full accounting in Appendix~\ref{app:workload}). We use BFCL's single-turn
categories because our accept unit is one drafted call scored against one target call;
\S\ref{sec:sealtools} repeats the protocol on Seal-Tools.

\paragraph{Models.} All experiments serve \texttt{gpt-oss-120b} \citep{openai2025gptoss},
an open-weight, natively tool-calling mixture-of-experts model (${\sim}5$B active of
$120$B total parameters), served with sglang on one NVIDIA H200 behind an
OpenAI-compatible endpoint; query embeddings use \texttt{all-MiniLM-L6-v2}. Each
unique query is answered once by the served model with greedy decoding. Compute
budget, licenses, and package versions are given in Appendix~\ref{app:reproducibility}.
We will release the harness, stream generator, and raw result files.

\paragraph{Baselines.} Two draft sources serve as baselines, sharing \method's target
model, embedder, and query stream; we refer to the three compared policies as
\emph{arms}. \emph{No memory} drafts the schema-only skeleton, the lower bound.
\emph{Static datastore} (\textsc{ToolSpec}-style) retrieves by cosine similarity from one global
store populated during warmup and then frozen (it never ingests and never evicts),
matching \textsc{ToolSpec}'s evaluation setting.

\paragraph{Metrics.} The payoff of a draft is its longest common token prefix
with what the target would emit (\S\ref{sec:method-prelim}). We decode the target's
tool-call stream once per query (arm-independent) and score each arm's draft by
token-level LCP, reporting \textbf{mean accepted tokens (MAT)}, the \textbf{accepted
fraction} of the target stream, and the \textbf{exact rate} (draft identical to
target). Calls are canonicalized (sorted keys) before scoring, so acceptance reflects
semantic agreement rather than wire-format noise. MAT is a \emph{controlled proxy},
distinct from two related measurements: (a) wire-format token acceptance in a deployed
decoder, which additionally depends on a trained drafter's distribution, tokenization,
and acceptance sampling (\S\ref{sec:limitations}); and (b) the wall-clock speedups of
the real external speculative loop in \S\ref{sec:maintable}, reported separately. MAT
is always measured against the served model's greedy
output; the arms differ only in the draft they retrieve.

\subsection{Acceptance results}

We first examine whether a continuously updated store improves acceptance after warmup
(Table~\ref{tab:accept}). \method\ reaches $19.96$ MAT,
compared with $15.70$ for the frozen store and $5.92$ for the no-memory baseline,
whose schema-only draft carries no history. The gain over the frozen store is $+4.26$ MAT
($+27\%$ relative), with a higher accepted fraction ($0.621$ vs.\ $0.492$) and exact-match
rate ($0.444$ vs.\ $0.314$). A paired bootstrap over the seed-0 replay places the absolute
gap at $+3.73$ MAT ($95\%$ CI $[3.37,4.10]$); the gap is positive for every user and every
session, and negative for only $5.8\%$ of recurring tasks.

\begin{table*}[t]
\centering
\caption{Post-warmup acceptance (sessions 1--11, $3$ seeds; per-arm $n$: BFCL $6{,}814$,
Seal-Tools $7{,}592$). Metrics as defined in \S\ref{sec:setup}; exact rate $=$ draft
identical to target.}
\label{tab:accept}
\small
\setlength{\tabcolsep}{4pt}
\begin{tabular}{llcccc}
\toprule
Benchmark & Arm & MAT & MAT seed-std & Accepted frac & Exact rate \\
\midrule
BFCL       & No memory (schema draft)        & 5.92  & $\pm0.07$ & 0.190 & 0.00 \\
           & Static datastore (frozen)       & 15.70 & $\pm0.20$ & 0.492 & 0.314 \\
           & \textbf{Ours (local, live, evict)} & \textbf{19.96} & $\pm0.05$ & \textbf{0.621} & \textbf{0.444} \\
\midrule
Seal-Tools & No memory (schema draft)        & 4.60  & $\pm0.06$ & 0.176 & 0.007 \\
           & Static datastore (frozen)       & 15.68 & $\pm0.27$ & 0.568 & 0.435 \\
           & \textbf{Ours (local, live, evict)} & \textbf{21.04} & $\pm0.37$ & \textbf{0.753} & \textbf{0.640} \\
\bottomrule
\end{tabular}
\end{table*}

\begin{figure}[t]
\centering
\includegraphics[width=\columnwidth]{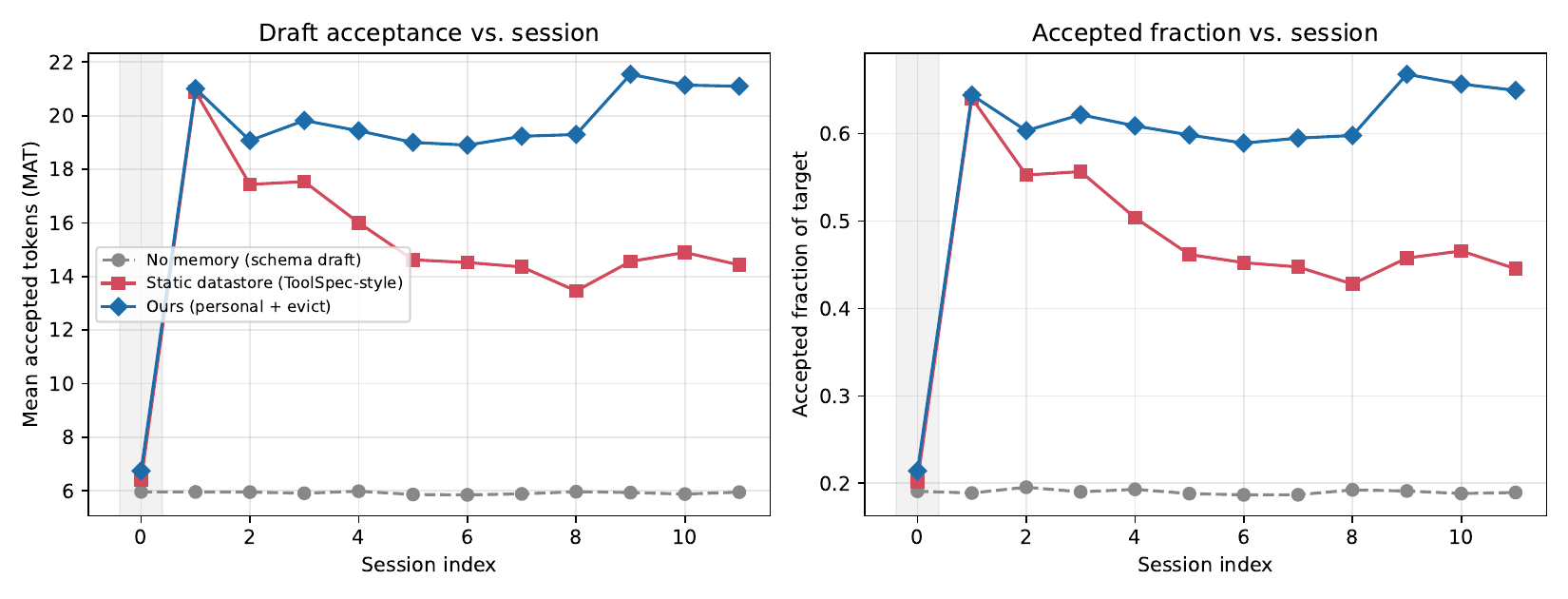}
\caption{Per-session MAT on the BFCL stream (3 seeds). The frozen store and \method\
are nearly tied at session~1; the frozen store then degrades while \method\ holds and the
no-memory floor stays flat.}
\label{fig:bfcl-sessions}
\end{figure}

The aggregate numbers alone do not capture the temporal behavior, which
Figure~\ref{fig:bfcl-sessions} reports per session.
At session~1 the two stores are nearly tied, both close to $21$ MAT. The frozen store then
degrades into a $13.5$--$14.9$ band: it can only replay its warmup snapshot, while later
sessions perturb the very argument tokens it drafts. \method\ instead holds at
$18.9$--$21.5$ by ingesting each session's verified calls, and the gap widens to $+46\%$
by the final session. The no-memory baseline stays flat near $5.9$ throughout. The one
non-monotonicity, a rise to $21.5$ at session~9, reflects the novel-task schedule rather
than noise. Figure~\ref{fig:freshness} shows the same dynamics on $\tau^2$-bench. The
frozen store's deficit therefore grows over time rather than remaining constant.

\subsection{Wall-clock results}
\label{sec:maintable}

\begin{figure}[t]
\centering
\includegraphics[width=\columnwidth]{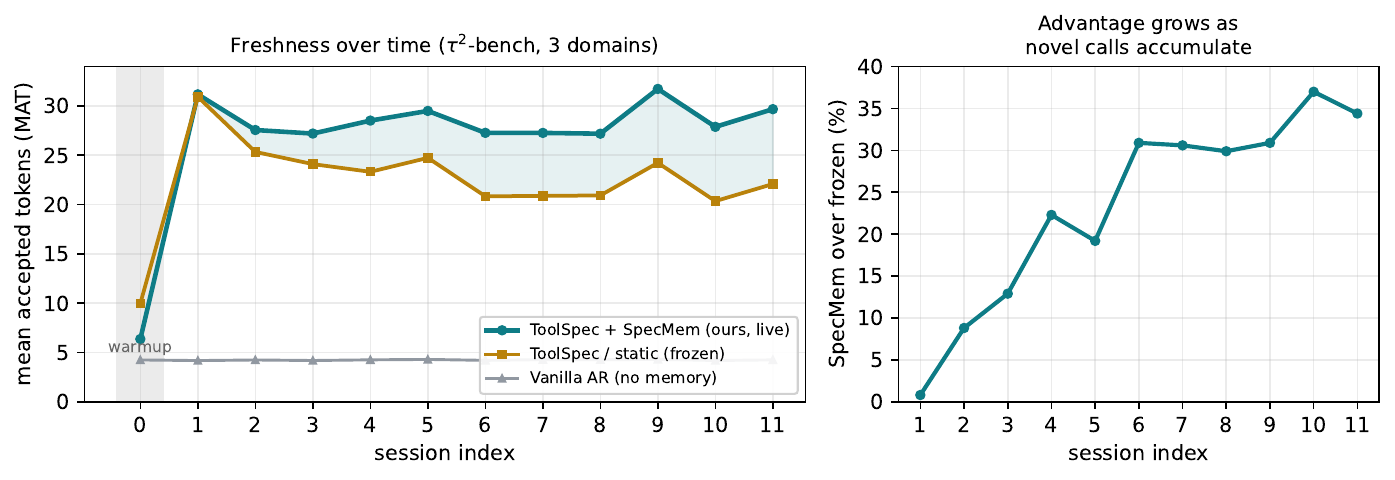}
\caption{Per-session MAT on $\tau^2$-bench ($3$ domains, $3$ seeds). At session~1 both
stores tie near $31$ MAT; the frozen \textsc{ToolSpec}/static store then degrades into a
$20$--$25$ band while \method\ holds at $27$--$32$. \textbf{Right:} the advantage over
the frozen store grows from ${\sim}1\%$ (session~1) to a peak of ${\sim}37\%$
(session~10; $34\%$ at the final session~11) as novel per-user calls accumulate, an effect a single-session benchmark cannot expose.}
\label{fig:freshness}
\end{figure}

We now compare against \textsc{ToolSpec} on its own four benchmarks (API-Bank
\citep{li2023apibank}, ToolAlpaca \citep{tang2023toolalpaca}, BFCL v4 \citep{bfcl2024},
and ToolBench \citep{qin2023toolllm}): to control a fair comparison, we reimplement it
with its components enabled (confidence-gated retrieval and the schema-aware FSM) inside
our harness,\footnote{Official code: \url{https://github.com/hemingkx/ToolSpec};
substitutions are listed in Appendix~\ref{app:toolspec}.} so every arm shares the same
served models, serving stack, and per-user session protocol. Speedup is real wall-clock
over vanilla autoregressive decoding, measured with an external speculative decoding
loop using serving-stack timers.

\begin{table}[t]
\centering
\caption{Wall-clock decode speedup vs.\ vanilla autoregressive decoding (p50) on
\textsc{ToolSpec}'s four benchmarks. Bottom rows: MAT for \method\ and for a live
token-match store at the same footprint ($3$ seeds).}
\label{tab:maintable}
\footnotesize
\setlength{\tabcolsep}{3pt}
\resizebox{\columnwidth}{!}{%
\begin{tabular}{lcccc}
\toprule
Method & API-Bank & ToolAlp. & BFCLv4 & ToolBench \\
\midrule
Vanilla AR          & $1.00\times$ & $1.00\times$ & $1.00\times$ & $1.00\times$ \\
\textsc{ToolSpec} (reimpl.) & $1.21\times$ & $1.66\times$ & $1.06\times$ & $1.58\times$ \\
\textbf{$+$ \method\ (ours)} & $\mathbf{1.62\times}$ & $\mathbf{1.74\times}$ & $\mathbf{1.18\times}$ & $\mathbf{1.69\times}$ \\
\midrule
\method\ MAT & $19.1$ & $20.9$ & $20.5$ & $21.0$ \\
Live token-match MAT & $19.1$ & $20.4$ & $19.8$ & $20.7$ \\
\bottomrule
\end{tabular}}
\end{table}

The reimplementation and the static arm of \S\ref{sec:setup} yield identical speedups
(MAT within $0.04$): the confidence gate and FSM rarely engage when retrieval is
reliable, so the gap to \method\ in Table~\ref{tab:maintable} is not a weak-baseline
artifact. Our reimplementation's speedups ($1.06$--$1.66\times$) are lower than the
$3.5$--$4.2\times$ reported on \textsc{ToolSpec}'s own stack, as expected under the
substitution: different served models and serving stack, an \emph{external} speculative
loop whose per-step client overhead bounds the achievable speedup, and BFCL v4 rather
than v2. Absolute speedups are therefore not comparable across papers; the within-table
contrasts, which hold these factors fixed, are.

Two observations complete the comparison. First, on standard single-session benchmarks
\method\ never regresses the reimplemented \textsc{ToolSpec} speedup, matching or
exceeding it on all four (Table~\ref{tab:maintable}); the ${\sim}9.4/9.5$ ms per query
that persistence spends on retrieval and write-back is included in the timed loop, so
persistence incurs no measurable penalty even where it provides no benefit, and the
advantage persists under concurrent load
(effective output throughput over vanilla decoding rises from $2.9\times$ to $3.3\times$
as concurrency grows from $1$ to $32$). Second, on
multi-session $\tau^2$-bench (\S\ref{sec:sealtools}) the frozen store's disadvantage
grows as novel calls accumulate (Figure~\ref{fig:freshness}), which single-session
benchmarks cannot expose.

\section{Analysis}
\label{sec:analysis}

\subsection{Ablation: attributing the gain}
\label{sec:ablation}

\method\ differs from the frozen store in three respects: it remains \emph{live}
(continuing to ingest after warmup), it scopes the store \emph{locally} per user, and it
\emph{evicts}. The aggregate gap alone cannot attribute the gain among these factors, so
we toggle each independently across four arms replayed on a shared target stream, using
the same three stream seeds as the headline run (Table~\ref{tab:ablation}). The live
\emph{global} arm serves as an attribution control rather than a deployment candidate:
it separates liveness from store scope and measures the cross-user leakage of a shared
store.\footnote{This separately run ablation regenerated its own
greedy targets, so its rows differ from Table~\ref{tab:accept}'s by at most $0.06$ MAT
(batched decoding is not bit-reproducible). Every contrast is within one run, so the
attribution is unaffected.}

\begin{table}[t]
\centering
\caption{Ablation: post-warmup MAT as each lifecycle component is toggled (full scale,
mean${\pm}$std over $3$ seeds).}
\label{tab:ablation}
\small
\setlength{\tabcolsep}{2pt}
\begin{tabular}{lcccc}
\toprule
Arm & live? & per-user? & evicts? & MAT \\
\midrule
Static (frozen)      & --- & --- & --- & $15.66{\scriptstyle\pm.11}$ \\
Global, live+evict   & \checkmark & --- & \checkmark & $20.06{\scriptstyle\pm.35}$ \\
Local, live, unbnd   & \checkmark & \checkmark & --- & $19.96{\scriptstyle\pm.29}$ \\
\textbf{Ours} & \checkmark & \checkmark & \checkmark & $\mathbf{19.91}{\scriptstyle\pm.29}$ \\
\bottomrule
\end{tabular}
\end{table}

The decomposition is stable across all three seeds: keeping
the store live accounts for essentially the entire gain. A \emph{global} store that keeps
growing and evicting but is shared by all users, at the same total footprint, already
reaches $20.06$ MAT, covering the full gap to the frozen store. Adding per-user
partitioning then changes MAT by $-0.15$ and eviction by $-0.05$, both negative in every
seed. Nor is the gain an artifact of the retriever: an arm that replaces our embedding
retrieval with \textsc{SuffixDecoding}-style exact token matching recovers it as well
(Table~\ref{tab:maintable}, bottom rows).

Note that the full configuration is therefore not the acceptance-optimal point: the live
global arm is slightly higher ($20.06$ vs.\ $19.91$). We nevertheless recommend the
local, evicting configuration: a global store serves one user drafts from another
user's history, and local scoping obtains isolation at a cost of only ${\sim}0.15$ MAT
(quantified below).

Although local scoping and eviction do not improve acceptance, they are what make a live
store deployable. Eviction bounds growth: a capacity sweep
shows that the headline capacity of $48$ costs only $0.04$--$0.05$ MAT relative to an
unbounded store, and acceptance degrades materially only below a
capacity of about $32$ at this stream's ${\sim}74$ queries per user. Partitioning provides isolation: replaying the live \emph{global} store, we find that
$14$--$68\%$ of drafts served to a user were written by a \emph{different} user (rising
with task overlap), a cross-user draft-provenance rate that a deployment
keeping users' histories separate would want at zero (cf.\
\citealp{mohammadi2026ghost}). A per-user store removes this channel by construction, at
${\sim}0$ acceptance cost in our disjoint-user regime.
Local retrieval adds no acceptance in the headline stream because disjoint users
already route each query to their own calls, so we also constructed the regime where it
plausibly \emph{could} win: users sharing tasks but holding fixed, user-specific
argument realizations. Even there the live global store is marginally better (local
costs $1.3$--$7.2\%$ MAT as overlap rises). Embedding retrieval already routes most
queries to the right user's entries, so the case for partitioning rests on isolation,
not accuracy.

Six further sweeps confirm the picture. With argument
perturbation turned off, the frozen store still loses $29\%$ to novel tasks arriving
after the freeze. Five randomized data constructions preserve the direction, with
liveness carrying $100$--$104\%$ of the gap. Session-reset and TTL arms attribute about
three quarters of the win to genuine cross-session memory. A warmup-budget sweep shrinks
the gap to ${\sim}2\%$ once warmup covers every task, confirming that the decay
reflects post-warmup novelty rather than an inadequate warmup. We do not claim that real
query streams necessarily drift this way; our claim is that a frozen store degrades
\emph{when} post-warmup arrival and recurrence are present. Finally, the decomposition
replicates on \texttt{gemma-4-31B-it}, where the live global arm recovers $4.00$ of the
$4.04$ MAT gap, so the finding holds on both a mixture-of-experts and a dense model. In
summary, liveness accounts for the acceptance gain, while local scope and eviction
provide isolation and boundedness.

\subsection{Side-effect gating for speculative execution}
\label{sec:safety}

\begin{table*}[t]
\centering
\caption{Side-effect-gating simulation ($400$-call streams, $5$ seeds; setup in
\S\ref{sec:safety}). Weighted cost $=\sum\sigma$
over wrong irreversible speculative executions; rate $=$ per drafted unsafe call; safe
spec-exec $=$ executions causing no irreversible harm.}
\label{tab:safety}
\small
\setlength{\tabcolsep}{4.5pt}
\begin{tabular}{lccccc}
\toprule
Policy & Weighted cost & Bad actions & Rate/opp. & Worst sev. & Safe spec-exec \\
\midrule
Naive spec-exec              & 800 & 124 & 0.75 & 10 & 255 \\
Confidence gate              & 255 & 40  & 0.24 & 10 & 244 \\
\textbf{Side-effect gate (ours)}  & \textbf{0} & \textbf{0} & \textbf{0.00} & \textbf{0} & 214 \\
\bottomrule
\end{tabular}
\end{table*}

We next evaluate speculative \emph{execution} of tools with irreversible side effects:
executing a drafted call before verification completes commits its side effect even if
the draft is subsequently rejected, and we quantify this cost under three gating
policies. The experiment is reported once: the gate decides from each tool's declared
annotation, not the LLM's output, so its numbers are model-independent
(\S\ref{sec:generalization}). The registry
contains $14$ side-effecting tools marked unsafe to speculate, spanning money movement,
data loss, security, infrastructure, and communication, each with an author-assigned
ordinal severity $\sigma\in[1,10]$, plus $7$ safe tools ($\sigma{=}0$). Each $400$-call
stream mixes in unsafe calls at rate $0.45$; the draft is the user's last call to the
same tool, and is wrong exactly when the user now targets a different object. Results
average $5$ stream seeds.

Table~\ref{tab:safety} compares the three policies of \S\ref{sec:method-safety} by the
severity-weighted cost of being wrong. Naively executing every accepted draft incurs a
weighted cost of $800$ from $124$ wrong irreversible executions. A confidence
gate, which executes an unsafe draft only when retrieval confidence exceeds a
threshold, helps but does not solve the problem: at $\tau_c{=}0.7$ it cuts the cost to
$255$, yet still commits $40$ bad actions, including maximum-severity ones. The reason is that retrieval confidence and side-effect safety are orthogonal: a stale
draft can remain highly similar to the query yet be wrong. Our primary evidence for this
is measured, not simulated: on real BFCL replays, wrong drafts occur at
retrieval similarity up to $0.9995$, indistinguishable from correct drafts at the
ceiling, so a threshold of $0.9$ still fires $33$--$60\%$ of wrong drafts, and the
threshold that reaches zero harm is a stream-dependent tail value unknowable in
advance. The side-effect gate's zero cost, by contrast, holds by
construction (refusing to execute unsafe tools trivially commits no harm), so the
experiment's value is quantifying the \emph{price} of that refusal against the harm
confidence gating admits: the gate defers the ${\sim}41$ correct unsafe drafts per stream
to post-verification execution while keeping all ${\sim}214$ safe speculations. The
policy ranking is unchanged under a raw error count ($124/40/0$) and a three-tier
severity map ($243/78/0$), so it is not an artifact of the $[1,10]$ weights. These
results indicate that confidence-based gating reduces but does not eliminate harm,
whereas gating on declared side effects eliminates it by construction.

\subsection{Generalization across served models}
\label{sec:generalization}

A single target model cannot tell us whether the gap reflects the memory policy or an
artifact of \texttt{gpt-oss-120b}. We therefore repeat the identical full-scale
protocol (same harness, stream, arms, and seeds) on two additional open-weight,
natively tool-calling models, each measured against its own greedy output with its own
tokenizer: \texttt{gemma-4-31B-it}, a dense $30.7$B model served with sglang, and
\texttt{Nemotron-3-Super-120B}, a hybrid-SSM model ($12$B active of $120$B total) served
with vLLM. The three models span vendor, scale, architecture, quantization, and tool-call
serialization format.

Per-model gating numbers are not re-measured: the side-effect gate keys off each tool's
declared annotation rather than the LLM that drafted the call, so its numbers are
identical across served models by construction.

Every model reproduces the same holds-versus-degrades dynamics, with a
$+27$--$29\%$ post-warmup gap and a $+45$--$48\%$ final-session gap.
Decline rates differ by model (${\sim}18\%$/$8\%$/$2\%$
of unique queries for \texttt{gpt-oss}/Nemotron/\texttt{gemma}); each model's declined
queries are excluded from its arms symmetrically. Since the gap reproduces
within two points across models that share none of these axes, we conclude it is a
property of the memory lifecycle rather than of any one target model.

\subsection{Seal-Tools and live multi-turn $\tau^2$-bench traces}
\label{sec:sealtools}

Cross-model replication reuses the same stream, so it cannot rule out an artifact of
the BFCL-derived pool. We therefore repeat the identical protocol (same construction,
scale, arms, seeds, and headline model) on Seal-Tools \citep{wu2024sealtools}, a
self-instruct function-calling benchmark ($700$-task test split, $4{,}076$ tools)
constructed independently of BFCL. Only about half of its queries contain perturbable
numerals, so recurrences are dominated by exact repeats. The result replicates and is slightly stronger
(Table~\ref{tab:accept}, bottom): the frozen store degrades from $22.6$ MAT into a
$13.2$--$14.6$ band while \method\ holds at $19.9$--$22.3$, a $+34\%$ post-warmup gap and
$+54\%$ in the final session. Exact repeats help both stores at warmup but do not
prevent the frozen store's decay from post-warmup task arrival.

\paragraph{$\tau^2$-bench.} To leave the single-shot regime entirely, we run the full
$278$-task $\tau^2$-bench set \citep{barres2025tau2} \emph{live} across the airline,
retail, and telecom domains, with the served \texttt{gpt-oss-120b} as the agent and a
live GPT-4.1 user simulator, so both states and targets are on-policy; the resulting
$1{,}753$ decision points feed the identical protocol over $3$ seeds. The frozen store reaches
$23.44{\scriptstyle\pm1.02}$ MAT and \method\ $28.64{\scriptstyle\pm.69}$, a $+22\%$ gap
overall. The per-domain gaps track each domain's share of novel calls, as the ablation
predicts: airline and retail, where most calls are unique
($82\%/56\%$), give $+27\%$, matching BFCL, while telecom, dominated by one shared
diagnostic workflow ($14\%$ unique calls), gives $+11\%$; when little novel work
arrives, a frozen store remains adequate.

\section{Conclusion}

We studied how persistent tool-call memory for speculative agents should be managed in
long-lived, multi-user deployments: online write-back sustains acceptance as the
stream evolves, local scope and eviction provide isolation and boundedness, and
future systems should keep draft stores live and gate speculative execution on declared
side effects, not retrieval confidence.

\section*{Limitations}
\label{sec:limitations}

\paragraph{Simulated repetition structure.} Our users and sessions are synthetic:
benchmark tasks partitioned across simulated users with numeric-variant perturbation,
not organic logs. The gap's \emph{magnitude} therefore depends on how much users repeat
their tools and how much novel work arrives after warmup, a dependence we measure
directly (\S\ref{sec:ablation}); its direction is
robust across seeds, drift rates, data constructions, and benchmarks. We surveyed
public candidates for real per-user, multi-session tool-call traces (LMSYS-Chat-1M,
WildChat, ToolBench, API-Bank, and the TraceLab agent traces) and none provide per-user
session structure together with the query and argument content our accept metric needs;
un-redacted multi-session agent telemetry remains the decisive test.

\paragraph{Metric and integration scope.} MAT scores one retrieved draft per query
against the served model's greedy output over canonicalized calls; it omits the
multi-turn-selection and irrelevance categories some benchmarks also test, and absolute
values are not comparable to trained drafters \citep{li2024eagle}. Wall-clock speedups
use a faithful external speculative loop with real serving-stack timers rather than an
engine-integrated hook. The retrieval threshold $\tau{=}0.35$ and the single embedder are
fixed, not swept, and the dynamics are confirmed on three open-weight model families but
not on deployed proprietary agents.

\paragraph{Side-effect gate scope.} The gate relies on correct per-tool \texttt{safe\_to\_speculate} annotations and
treats side-effect safety as binary; on call-dependent tools a per-call ternary gate strictly
dominates and is the natural refinement. Rollback, compensating actions, and
learned idempotency inference are open directions.

\section*{Ethical Considerations}

Persistent memory and speculative execution for tool-calling agents introduce privacy and operational safety risks. A globally shared memory may expose one user's interaction history through drafts retrieved for another user, motivating our use of per-user memory partitioning. Incorrect speculative execution may also trigger irreversible actions before verification completes; \method therefore defers tools declared unsafe to speculate until verification is complete. These protections still depend on correct tool annotations and should be complemented in deployment by access control, auditing, sandboxing, and human confirmation for high-impact actions. Our experiments use public benchmarks and simulated users rather than real user logs.

\bibliography{refs}

@article{xia2026toolspec,
  title   = {{ToolSpec}: Accelerating Tool Calling via Schema-Aware and
             Retrieval-Augmented Speculative Decoding},
  author  = {Xia, Heming and Li, Yongqi and Du, Cunxiao and Song, Mingbo and Li, Wenjie},
  journal = {arXiv preprint arXiv:2604.13519},
  note    = {arXiv preprint (v1 April 2026, v2 May 2026)},
  year    = {2026},
  eprint  = {2604.13519},
  version = {v2},
  archivePrefix = {arXiv},
  primaryClass  = {cs.CL}
}

@article{sui2026paste,
  title={Parallelizing Tool Execution and LLM Generation for Low-Latency Agent Serving},
  author={Sui, Yifan and Zhao, Han and Ma, Rui and He, Zhiyuan and Wang, Hao and Li, Jianxun and Xu, Kaiqiang and Chen, Kai and Yang, Yuqing},
  journal={arXiv preprint arXiv:2603.18897},
  year={2026}
}

@article{saberi2026spechop,
  title   = {{SpecHop}: Continuous Speculation for Accelerating Multi-Hop
             Retrieval Agents},
  author  = {Saberi, Mehrdad and Rezaei, Keivan and Feizi, Soheil},
  journal = {arXiv preprint arXiv:2605.21965},
  year    = {2026},
  eprint  = {2605.21965},
  archivePrefix = {arXiv},
  primaryClass  = {cs.CL}
}

@article{chung2026agentx,
  title   = {{Agent-X}: Full Pipeline Acceleration of On-device {AI} Agents},
  author  = {Chung, Jinha and Shin, Byeongjun and Kim, Jiin and Rhu, Minsoo},
  journal = {arXiv preprint arXiv:2605.10380},
  year    = {2026},
  eprint  = {2605.10380},
  archivePrefix = {arXiv},
  primaryClass  = {cs.LG},
  note = {arXiv:2605.10380; accepted at MobiSys 2026}
}

@article{hooper2026specagents,
  title   = {Speculative Interaction Agents: Building Real-Time Agents with
             Asynchronous {I/O} and Speculative Tool Calling},
  author  = {Hooper, Coleman and Kang, Minwoo and Moon, Suhong and Lee, Nicholas and
             Wen, Eric and Wawrzynek, John and Mahoney, Michael W. and
             Shao, Yakun Sophia and Gholami, Amir and Keutzer, Kurt},
  journal = {arXiv preprint arXiv:2605.13360},
  year    = {2026},
  eprint  = {2605.13360},
  archivePrefix = {arXiv},
  primaryClass  = {cs.CL}
}

@inproceedings{oliaro2024suffixdecoding,
  title   = {{SuffixDecoding}: Extreme Speculative Decoding for Emerging AI Applications},
  author  = {Oliaro, Gabriele and Jia, Zhihao and Campos, Daniel and Qiao, Aurick},
  booktitle = {Advances in Neural Information Processing Systems (NeurIPS), Spotlight},
  year    = {2025},
  eprint  = {2411.04975},
  archivePrefix = {arXiv},
  primaryClass  = {cs.CL}
}

@inproceedings{he2024rest,
  title     = {{REST}: Retrieval-Based Speculative Decoding},
  author    = {He, Zhenyu and Zhong, Zexuan and Cai, Tianle and Lee, Jason D. and He, Di},
  booktitle = {Proceedings of the 2024 Conference of the North American Chapter of the
               Association for Computational Linguistics (NAACL)},
  year      = {2024},
  note      = {arXiv:2311.08252},
  eprint    = {2311.08252},
  archivePrefix = {arXiv},
  primaryClass  = {cs.CL}
}

@inproceedings{leviathan2023speculative,
  title     = {Fast Inference from Transformers via Speculative Decoding},
  author    = {Leviathan, Yaniv and Kalman, Matan and Matias, Yossi},
  booktitle = {Proceedings of the 40th International Conference on Machine Learning (ICML)},
  year      = {2023},
  eprint    = {2211.17192},
  archivePrefix = {arXiv},
  primaryClass  = {cs.LG}
}

@article{chen2023accelerating,
  title   = {Accelerating Large Language Model Decoding with Speculative Sampling},
  author  = {Chen, Charlie and Borgeaud, Sebastian and Irving, Geoffrey and
             Lespiau, Jean-Baptiste and Sifre, Laurent and Jumper, John},
  journal = {arXiv preprint arXiv:2302.01318},
  year    = {2023},
  eprint  = {2302.01318},
  archivePrefix = {arXiv},
  primaryClass  = {cs.CL}
}

@inproceedings{cai2024medusa,
  title     = {{Medusa}: Simple {LLM} Inference Acceleration Framework with Multiple
               Decoding Heads},
  author    = {Cai, Tianle and Li, Yuhong and Geng, Zhengyang and Peng, Hongwu and
               Lee, Jason D. and Chen, Deming and Dao, Tri},
  booktitle = {Proceedings of the 41st International Conference on Machine Learning (ICML)},
  year      = {2024},
  eprint    = {2401.10774},
  archivePrefix = {arXiv},
  primaryClass  = {cs.LG}
}

@inproceedings{li2024eagle,
  title     = {{EAGLE}: Speculative Sampling Requires Rethinking Feature Uncertainty},
  author    = {Li, Yuhui and Wei, Fangyun and Zhang, Chao and Zhang, Hongyang},
  booktitle = {Proceedings of the 41st International Conference on Machine Learning (ICML)},
  year      = {2024},
  eprint    = {2401.15077},
  archivePrefix = {arXiv},
  primaryClass  = {cs.CL}
}

@inproceedings{fu2024lookahead,
  title     = {Break the Sequential Dependency of {LLM} Inference Using Lookahead Decoding},
  author    = {Fu, Yichao and Bailis, Peter and Stoica, Ion and Zhang, Hao},
  booktitle = {Proceedings of the 41st International Conference on Machine Learning (ICML)},
  year      = {2024},
  eprint    = {2402.02057},
  archivePrefix = {arXiv},
  primaryClass  = {cs.LG}
}

@inproceedings{xia2024specbench,
    title = "Unlocking Efficiency in Large Language Model Inference: A Comprehensive Survey of Speculative Decoding",
    author = "Xia, Heming  and
      Yang, Zhe  and
      Dong, Qingxiu  and
      Wang, Peiyi  and
      Li, Yongqi  and
      Ge, Tao  and
      Liu, Tianyu  and
      Li, Wenjie  and
      Sui, Zhifang",
    editor = "Ku, Lun-Wei  and
      Martins, Andre  and
      Srikumar, Vivek",
    booktitle = "Findings of the Association for Computational Linguistics: ACL 2024",
    month = aug,
    year = "2024",
    address = "Bangkok, Thailand",
    publisher = "Association for Computational Linguistics",
    url = "https://aclanthology.org/2024.findings-acl.456/",
    doi = "10.18653/v1/2024.findings-acl.456",
    pages = "7655--7671",
}

@misc{bfcl2024,
  title        = {Berkeley Function Calling Leaderboard},
  author       = {Yan, Fanjia and Mao, Huanzhi and Ji, Charlie Cheng-Jie and
                  Zhang, Tianjun and Patil, Shishir G. and Stoica, Ion and
                  Gonzalez, Joseph E.},
  year         = {2024},
  howpublished = {\url{https://gorilla.cs.berkeley.edu/blogs/8_berkeley_function_calling_leaderboard.html}}
}

@misc{openai2025gptoss,
  title        = {{gpt-oss-120b} and {gpt-oss-20b} Model Card},
  author       = {{OpenAI}},
  year         = {2025},
  howpublished = {\url{https://huggingface.co/openai/gpt-oss-120b}}
}

@article{li2024eagle2,
  title   = {{EAGLE-2}: Faster Inference of Language Models with Dynamic Draft Trees},
  author  = {Li, Yuhui and Wei, Fangyun and Zhang, Chao and Zhang, Hongyang},
  journal = {arXiv preprint arXiv:2406.16858},
  year    = {2024},
  eprint  = {2406.16858},
  archivePrefix = {arXiv},
  primaryClass  = {cs.CL}
}

@article{yao2022react,
  title   = {{ReAct}: Synergizing Reasoning and Acting in Language Models},
  author  = {Yao, Shunyu and Zhao, Jeffrey and Yu, Dian and Du, Nan and
             Shafran, Izhak and Narasimhan, Karthik and Cao, Yuan},
  journal = {arXiv preprint arXiv:2210.03629},
  year    = {2022},
  eprint  = {2210.03629},
  archivePrefix = {arXiv},
  primaryClass  = {cs.CL}
}

@article{schick2023toolformer,
  title   = {{Toolformer}: Language Models Can Teach Themselves to Use Tools},
  author  = {Schick, Timo and Dwivedi-Yu, Jane and Dess{\`i}, Roberto and
             Raileanu, Roberta and Lomeli, Maria and Zettlemoyer, Luke and
             Cancedda, Nicola and Scialom, Thomas},
  journal = {arXiv preprint arXiv:2302.04761},
  year    = {2023},
  eprint  = {2302.04761},
  archivePrefix = {arXiv},
  primaryClass  = {cs.CL}
}

@article{qin2023toolllm,
  title   = {{ToolLLM}: Facilitating Large Language Models to Master
             16000+ Real-world {APIs}},
  author  = {Qin, Yujia and Liang, Shihao and Ye, Yining and Zhu, Kunlun and
             Yan, Lan and Lu, Yaxi and Lin, Yankai and Cong, Xin and Tang, Xiangru
             and Qian, Bill and others},
  journal = {arXiv preprint arXiv:2307.16789},
  year    = {2023},
  eprint  = {2307.16789},
  archivePrefix = {arXiv},
  primaryClass  = {cs.AI}
}

@article{tang2023toolalpaca,
  title   = {{ToolAlpaca}: Generalized Tool Learning for Language Models with 3000 Simulated Cases},
  author  = {Tang, Qiaoyu and Deng, Ziliang and Lin, Hongyu and Han, Xianpei and
             Liang, Qiao and Cao, Boxi and Sun, Le},
  journal = {arXiv preprint arXiv:2306.05301},
  year    = {2023},
}

@article{li2023apibank,
  title   = {{API-Bank}: A Comprehensive Benchmark for Tool-Augmented {LLMs}},
  author  = {Li, Minghao and Zhao, Yingxiu and Yu, Bowen and Song, Feifan and
             Li, Hangyu and Yu, Haiyang and Li, Zhoujun and Huang, Fei and Li, Yongbin},
  journal = {arXiv preprint arXiv:2304.08244},
  year    = {2023},
  eprint  = {2304.08244},
  archivePrefix = {arXiv},
  primaryClass  = {cs.CL}
}

@article{liu2024toolace,
  title   = {{ToolACE}: Winning the Points of {LLM} Function Calling},
  author  = {Liu, Weiwen and Huang, Xu and Zeng, Xingshan and Hao, Xinlong and
             Yu, Shuai and Li, Dexun and Wang, Shuai and Gan, Weinan and Liu, Zhengying
             and others},
  journal = {arXiv preprint arXiv:2409.00920},
  year    = {2024},
  eprint  = {2409.00920},
  archivePrefix = {arXiv},
  primaryClass  = {cs.LG}
}

@article{packer2023memgpt,
  title   = {{MemGPT}: Towards {LLMs} as Operating Systems},
  author  = {Packer, Charles and Wooders, Sarah and Lin, Kevin and Fang, Vivian and
             Patil, Shishir G. and Stoica, Ion and Gonzalez, Joseph E.},
  journal = {arXiv preprint arXiv:2310.08560},
  year    = {2023},
  eprint  = {2310.08560},
  archivePrefix = {arXiv},
  primaryClass  = {cs.AI}
}

@article{zhong2023memorybank,
  title   = {{MemoryBank}: Enhancing Large Language Models with Long-Term Memory},
  author  = {Zhong, Wanjun and Guo, Lianghong and Gao, Qiqi and Ye, He and Wang, Yanlin},
  journal = {arXiv preprint arXiv:2305.10250},
  year    = {2023},
  eprint  = {2305.10250},
  archivePrefix = {arXiv},
  primaryClass  = {cs.CL}
}

@article{park2023generative,
  title   = {Generative Agents: Interactive Simulacra of Human Behavior},
  author  = {Park, Joon Sung and O'Brien, Joseph C. and Cai, Carrie J. and
             Morris, Meredith Ringel and Liang, Percy and Bernstein, Michael S.},
  journal = {arXiv preprint arXiv:2304.03442},
  year    = {2023},
  eprint  = {2304.03442},
  archivePrefix = {arXiv},
  primaryClass  = {cs.HC}
}

@article{shinn2023reflexion,
  title   = {{Reflexion}: Language Agents with Verbal Reinforcement Learning},
  author  = {Shinn, Noah and Cassano, Federico and Berman, Edward and
             Gopinath, Ashwin and Narasimhan, Karthik and Yao, Shunyu},
  journal = {arXiv preprint arXiv:2303.11366},
  year    = {2023},
  eprint  = {2303.11366},
  archivePrefix = {arXiv},
  primaryClass  = {cs.AI}
}

@article{xu2025amem,
  title   = {{A-MEM}: Agentic Memory for {LLM} Agents},
  author  = {Xu, Wujiang and Liang, Zujie and Mei, Kai and Gao, Hang and
             Tan, Juntao and Zhang, Yongfeng},
  journal = {arXiv preprint arXiv:2502.12110},
  year    = {2025},
  eprint  = {2502.12110},
  archivePrefix = {arXiv},
  primaryClass  = {cs.CL}
}

@article{ruan2023toolemu,
  title   = {Identifying the Risks of {LM} Agents with an {LM}-Emulated Sandbox},
  author  = {Ruan, Yangjun and Dong, Honghua and Wang, Andrew and Pitis, Silviu and
             Zhou, Yongchao and Ba, Jimmy and Dubois, Yann and Maddison, Chris J.
             and Hashimoto, Tatsunori},
  journal = {arXiv preprint arXiv:2309.15817},
  year    = {2023},
  eprint  = {2309.15817},
  archivePrefix = {arXiv},
  primaryClass  = {cs.AI}
}

@article{yuan2024rjudge,
  title   = {{R-Judge}: Benchmarking Safety Risk Awareness for {LLM} Agents},
  author  = {Yuan, Tongxin and He, Zhiwei and Dong, Lingzhong and Wang, Yiming and
             Zhao, Ruijie and Xia, Tian and Xu, Lizhen and Zhou, Binglin and
             Li, Fangqi and Zhang, Zhuosheng and others},
  journal = {arXiv preprint arXiv:2401.10019},
  year    = {2024},
  eprint  = {2401.10019},
  archivePrefix = {arXiv},
  primaryClass  = {cs.CL}
}

@article{ye2024toolsword,
  title   = {{ToolSword}: Unveiling Safety Issues of Large Language Models in
             Tool Learning Across Three Stages},
  author  = {Ye, Junjie and Li, Sixian and Li, Guanyu and Huang, Caishuang and
             Gao, Songyang and Wu, Yilong and Zhang, Qi and Gui, Tao and Huang, Xuanjing},
  journal = {arXiv preprint arXiv:2402.10753},
  year    = {2024},
  eprint  = {2402.10753},
  archivePrefix = {arXiv},
  primaryClass  = {cs.CL}
}

@article{xiang2024guardagent,
  title   = {{GuardAgent}: Safeguard {LLM} Agents by a Guard Agent via
             Knowledge-Enabled Reasoning},
  author  = {Xiang, Zhen and Zheng, Linzhi and Li, Yanjie and Hong, Junyuan and
             Li, Qinbin and Xie, Han and Zhang, Jiawei and Xiong, Zidi and Xie, Chulin
             and others},
  journal = {arXiv preprint arXiv:2406.09187},
  year    = {2024},
  eprint  = {2406.09187},
  archivePrefix = {arXiv},
  primaryClass  = {cs.CR}
}

@article{bai2026spork,
  title   = {{SPORK}: Self-Speculative Forking to Accelerate Agentic {LLM} Inference},
  author  = {Bai, Huajun and Lv, Weiwei and Zheng, Huichuan and Lu, Youyou and Shu, Jiwu},
  journal = {arXiv preprint arXiv:2607.03333},
  year    = {2026},
  eprint  = {2607.03333},
  archivePrefix = {arXiv},
  primaryClass  = {cs.LG}
}

@article{song2026bpaste,
  title   = {{B-PASTE}: Beam-Aware Pattern-Guided Speculative Execution for
             Resource-Constrained {LLM} Agents},
  author  = {Song, Yanfei},
  journal = {arXiv preprint arXiv:2604.16469},
  year    = {2026},
  eprint  = {2604.16469},
  archivePrefix = {arXiv},
  primaryClass  = {cs.LG}
}

@article{mohammadi2026ghost,
  title   = {Ghost Tool Calls: Issue-Time Privacy for Speculative Agent Tools},
  author  = {Mohammadi, Bardia and Klein, Lars and Arora, Akhil and Bindschaedler, Laurent},
  journal = {arXiv preprint arXiv:2606.02483},
  year    = {2026},
  eprint  = {2606.02483},
  archivePrefix = {arXiv},
  primaryClass  = {cs.CR}
}

@article{huang2026speceyes,
  title   = {{SpecEyes}: Accelerating Agentic Multimodal {LLMs} via Speculative
             Perception and Planning},
  author  = {Huang, Haoyu and Huang, Jinfa and Wan, Zhongwei and Zheng, Xiawu and
             Ji, Rongrong and Luo, Jiebo},
  journal = {arXiv preprint arXiv:2603.23483},
  year    = {2026},
  eprint  = {2603.23483},
  archivePrefix = {arXiv},
  primaryClass  = {cs.CV},
  note = {arXiv:2603.23483; accepted at ECCV 2026}
}

@inproceedings{liu2024online,
  title     = {Online Speculative Decoding},
  author    = {Liu, Xiaoxuan and Hu, Lanxiang and Bailis, Peter and Cheung, Alvin and
               Deng, Zhijie and Stoica, Ion and Zhang, Hao},
  booktitle = {Proceedings of the 41st International Conference on Machine Learning (ICML)},
  year      = {2024},
  eprint    = {2310.07177},
  archivePrefix = {arXiv},
  primaryClass  = {cs.LG}
}

@article{wang2024awm,
  title   = {Agent Workflow Memory},
  author  = {Wang, Zora Zhiruo and Mao, Jiayuan and Fried, Daniel and Neubig, Graham},
  journal = {arXiv preprint arXiv:2409.07429},
  year    = {2024},
  eprint  = {2409.07429},
  archivePrefix = {arXiv},
  primaryClass  = {cs.CL}
}

@inproceedings{bang2023gptcache,
  title     = {{GPTCache}: An Open-Source Semantic Cache for {LLM} Applications
               Enabling Faster Answers and Cost Savings},
  author    = {Bang, Fu},
  booktitle = {Proceedings of the 3rd Workshop for Natural Language Processing
               Open Source Software (NLP-OSS)},
  year      = {2023},
  url       = {https://aclanthology.org/2023.nlposs-1.24}
}

@inproceedings{nightingale2005speculative,
  title     = {Speculative Execution in a Distributed File System},
  author    = {Nightingale, Edmund B. and Chen, Peter M. and Flinn, Jason},
  booktitle = {Proceedings of the 20th ACM Symposium on Operating Systems
               Principles (SOSP)},
  year      = {2005},
  doi       = {10.1145/1095810.1095829}
}

@article{wu2024sealtools,
  title   = {Seal-Tools: Self-Instruct Tool Learning Dataset for Agent Tuning and
             Detailed Benchmark},
  author  = {Wu, Mengsong and Zhu, Tong and Han, Han and Tan, Chuanyuan and
             Zhang, Xiang and Chen, Wenliang},
  journal = {arXiv preprint arXiv:2405.08355},
  year    = {2024},
  eprint  = {2405.08355},
  archivePrefix = {arXiv},
  primaryClass  = {cs.CL}
}

@article{barres2025tau2,
  title   = {$\tau^2$-Bench: Evaluating Conversational Agents in a Dual-Control Environment},
  author  = {Barres, Victor and Dong, Honghua and Ray, Soham and Si, Xujie and
             Narasimhan, Karthik},
  journal = {arXiv preprint arXiv:2506.07982},
  year    = {2025},
  eprint  = {2506.07982},
  archivePrefix = {arXiv},
  primaryClass  = {cs.AI}
}

\clearpage
\appendix
\section{Experimental Details}
\label{app:details}

\subsection{Stream details}
\label{app:workload}

\textbf{Query accounting.} Session~0 (warmup) issues each user's $8$ warmup tasks exactly
once; sessions~1--11 issue $6$ queries each, so one seed contributes
$40\times(8+11\times6)=2{,}960$ queries and three seeds give $8{,}880$. Exact repeats and
perturbation collisions leave $5{,}535$ unique query strings (BFCL; $5{,}088$ on Seal-Tools),
each answered once by the served model with greedy decoding; no target is fabricated. The
model declines a tool call on $982$ unique queries (BFCL; $238$ on Seal-Tools); those are
excluded from \emph{all} arms symmetrically, leaving $n{=}6{,}814$ (BFCL) and $n{=}7{,}592$
(Seal-Tools) scored post-warmup queries per arm. Session~0 is a cold start for every arm,
so all headline comparisons are on sessions~1--11. (\texttt{exact\_rate}${\approx}0$ for the
no-memory arm is expected: a schema-only draft matches a concrete call only when that call
takes no arguments (never on BFCL, $0.7\%$ on Seal-Tools).)

\textbf{Arrival schedule.} Warmup covers half of each user's task set ($8$ of $15$
tasks, \texttt{warmup\_frac}$=0.5$). Post-warmup sessions draw each user's queries
from their full task set with unwarmed (novel) tasks weighted $3{:}1$ over warmed
ones, spreading first appearances across sessions~1--11 (\texttt{arrival}$=$
\texttt{spread}); \texttt{burst} and \texttt{late} alternatives, and warmup fractions up
to $1.0$, were additionally swept (stable in direction throughout). The headline gap should be read at this
operating point: as the warmup sweep shows, it shrinks to ${\sim}2\%$ when warmup covers
every task.

\textbf{Scope of exclusions.} Declined queries are removed from all arms symmetrically,
so comparisons are internally valid but conclusions apply to the subpopulation of queries
on which the served model issues a tool call at all ($82\%$ of unique queries for
\texttt{gpt-oss} on BFCL, $95\%$ on Seal-Tools).

\textbf{$\tau^2$-bench trace generation.} For the live $\tau^2$-bench experiments
(\S\ref{sec:sealtools}), one canonical trace is generated per task and served model
(a single trial): the served model acts as the agent against a live GPT-4.1 user
simulator, and the $278$ tasks yield $1{,}753$ post-warmup decision points. All
memory arms then replay the same trace, so arms differ only in the draft they
retrieve; multiple trials would alter the success statistics of the underlying
benchmark but not this within-trace acceptance comparison. The seeds reported there
are stream-construction seeds for the replay, which reuses cached targets.

\textbf{Pool choice.} The disjoint-per-user design needs
$\text{users}\times\text{tasks}=600$ distinct examples, hence $600$ of BFCL's $800$-task
simple/multiple/parallel pool. We restrict to the single-turn categories because our accept
unit is one drafted call scored against one target call; BFCL's multi-turn and
irrelevance-detection categories evaluate dialogue-level tool selection, which this protocol
does not measure (\S\ref{sec:limitations}).

\subsection{\textsc{ToolSpec}-style reimplementation}
\label{app:toolspec}

We reimplement \textsc{ToolSpec} inside our harness rather than running its official
release (\url{https://github.com/hemingkx/ToolSpec}), which targets its own models and
serving stack; reimplementation lets every arm share our served models, session
protocol, and timers. We reproduce its two mechanisms over a fixed,
global store shared across users: (i) \emph{confidence-gated retrieval} (return the nearest
stored call only above a similarity threshold, vs.\ the static proxy that returns its
nearest neighbor unconditionally); (ii) a \emph{schema-aware FSM surrogate} (on a cold
miss, emit a structurally valid draft for the nearest function rather than a far wrong
call). This is provably at least as strong as the static proxy; on every stream here the two
frozen arms produce identical wall-clock speedups (MAT within $0.04$; exactly identical on
$\tau^2$-bench), so the gap to \method\ is not a weak-baseline artifact.
For Table~\ref{tab:maintable} we run each of \textsc{ToolSpec}'s own four benchmarks
(API-Bank, ToolAlpaca, BFCL v4, ToolBench from its OpenBMB release) through the identical
per-user/session construction ($10$ tasks/user, $\min(40,|\mathcal{D}|/10)$ users, $12$
sessions, $3$ seeds), with targets from the served \texttt{gpt-oss-120b}. Wall-clock uses the
external spec-decode loop (real sglang timers, \texttt{ignore\_eos} forcing exact
token counts). We use BFCL v4 (superset of the v2
\textsc{ToolSpec} used) and our own served models rather than \textsc{ToolSpec}'s
checkpoints, stated as a substitution rather than an exact reproduction; we therefore call this a reimplementation throughout. Retrieval/write-back cost
${\sim}9.4/9.5$ ms/query, inside the timed loop.

\section{Reproducibility Details}
\label{app:reproducibility}

\subsection{Artifact Licenses}
\label{app:licenses}

The models and benchmarks used in this work were released under licenses permitting research use. GPT-OSS-120B and Gemma-4-31B-IT are distributed under the Apache License 2.0, while Nemotron-3-Super-120B is governed by the NVIDIA Nemotron Open Model License. BFCL, Seal-Tools, ToolAlpaca, and ToolBench are distributed under the Apache License 2.0; API-Bank and $\tau^2$-bench are distributed under the MIT License. Our release will contain only our experimental harness, stream-construction code, configurations, and derived results. 
\subsection{Compute and Implementation Details}
\label{app:compute}

\paragraph{Models.}
We serve three open-weight, natively tool-calling models: \texttt{gpt-oss-120b}
(mixture-of-experts, $120$B total, ${\sim}5$B active parameters, MXFP4),
\texttt{gemma-4-31B-it} ($30.7$B, dense), and \texttt{Nemotron-3-Super-120B}
(hybrid-SSM, $120$B total, $12$B active). Query embeddings use
\texttt{all-MiniLM-L6-v2} ($22$M parameters), run on CPU. The live $\tau^2$-bench
user simulator is GPT-4.1, accessed through the OpenAI API.

\paragraph{Computing infrastructure and budget.}
Experiments ran on NVIDIA H200 GPUs: one GPU each for \texttt{gpt-oss-120b} and
\texttt{gemma-4-31B-it}, and four for \texttt{Nemotron-3-Super-120B} (tensor
parallelism $4$). The full study required approximately $20$ H200 GPU-hours of
active compute.

\paragraph{Hyperparameters.}
Targets are decoded greedily. \method\ uses top-$1$ cosine retrieval with threshold
$\tau=0.35$ and a per-user capacity of $48$ entries under LRU eviction, fixed in
pilot runs and held constant across all benchmarks and models; capacity is swept in
\S\ref{sec:ablation}. Stream-construction parameters are given in
\S\ref{sec:method-workload} and Appendix~\ref{app:workload}; all runs use stream
seeds $0$, $1$, $2$.

\paragraph{Software.}
The harness runs on Python $3.10$ with PyTorch $2.9.1$ and Transformers $5.6.0$;
models are served with sglang $0.5.11$ (\texttt{gpt-oss-120b},
\texttt{gemma-4-31B-it}) and vLLM $0.18$ (\texttt{Nemotron-3-Super-120B}) behind
OpenAI-compatible endpoints.

\end{document}